\documentclass[letterpaper, 10 pt, conference]{ieeeconf}  % Comment this

\IEEEoverridecommandlockouts                              % This command is only needed if 
\usepackage{multirow}
\usepackage{amsmath}  % 用于换行, 设置行间距
\usepackage{amssymb}  % 用来画勾勾
\usepackage{color,array}
\usepackage{xcolor}
\usepackage{graphicx}
\usepackage{setspace}  % 设置行间距
\usepackage{makecell}
\usepackage{subcaption}
\usepackage{caption}
\DeclareCaptionLabelSeparator{periodspace}{.\quad}
\usepackage{graphics} % for pdf, bitmapped graphics files
\usepackage{epsfig} % for postscript graphics files
\usepackage{times} % assumes new font selection scheme installed
\usepackage{amsmath} % assumes amsmath package installed
\usepackage{amssymb}  % assumes amsmath package installed
\usepackage[table]{xcolor}

\usepackage[linesnumbered,ruled,vlined]{algorithm2e}
\usepackage{subfloat}

\usepackage{booktabs} %用三线表
\usepackage{gensymb}  % 用来表示度数°
\usepackage{bm}  % 用来给希腊字母加粗
\usepackage{underscore}  % 使用下划线

\usepackage{amsmath} % assumes amsmath package installed
\usepackage{amssymb}  % assumes amsmath package installed

\usepackage{threeparttable}
\usepackage{enumerate}
\usepackage{bbm}

\usepackage[hidelinks]{hyperref}

\usepackage{bbding} %首先在导言区调用bbding包
\usepackage{cite}
\usepackage{hyperref}

\begin{document}
\title{\LARGE \bf
TARIC: Memory-Augmented Traversability-Aware Outdoor VLN under Interrupted Semantic Cues
}

\author{
Tianle Zeng$^{1}$, Hanjing Ye$^{1}$, Jianwei Peng$^{1}$, Jingwen Yu$^{1,2}$, Hanxuan Chen$^{3}$, and Hong Zhang$^{1,*}$%
\thanks{$^{1}$ Shenzhen Key Laboratory of Robotics and Computer Vision, Southern University of Science and Technology, Shenzhen, China.}%
\thanks{$^{2}$
        CKS Robotics Institute, Hong Kong University of Science and Technology, Hong Kong SAR, China.}
\thanks{$^{3}$ College of Electrical and Information Engineering, Hunan University, Hunan, China.}
\thanks{$^{*}$Corresponding author: Hong Zhang. Contact email: \texttt{hzhang@sustech.edu.cn}.}
}

\makeatletter
\let\@oldmaketitle\@maketitle%
\renewcommand{\@maketitle}{\@oldmaketitle%
    \centering
    \vspace*{1mm}
    \includegraphics[width=\textwidth]{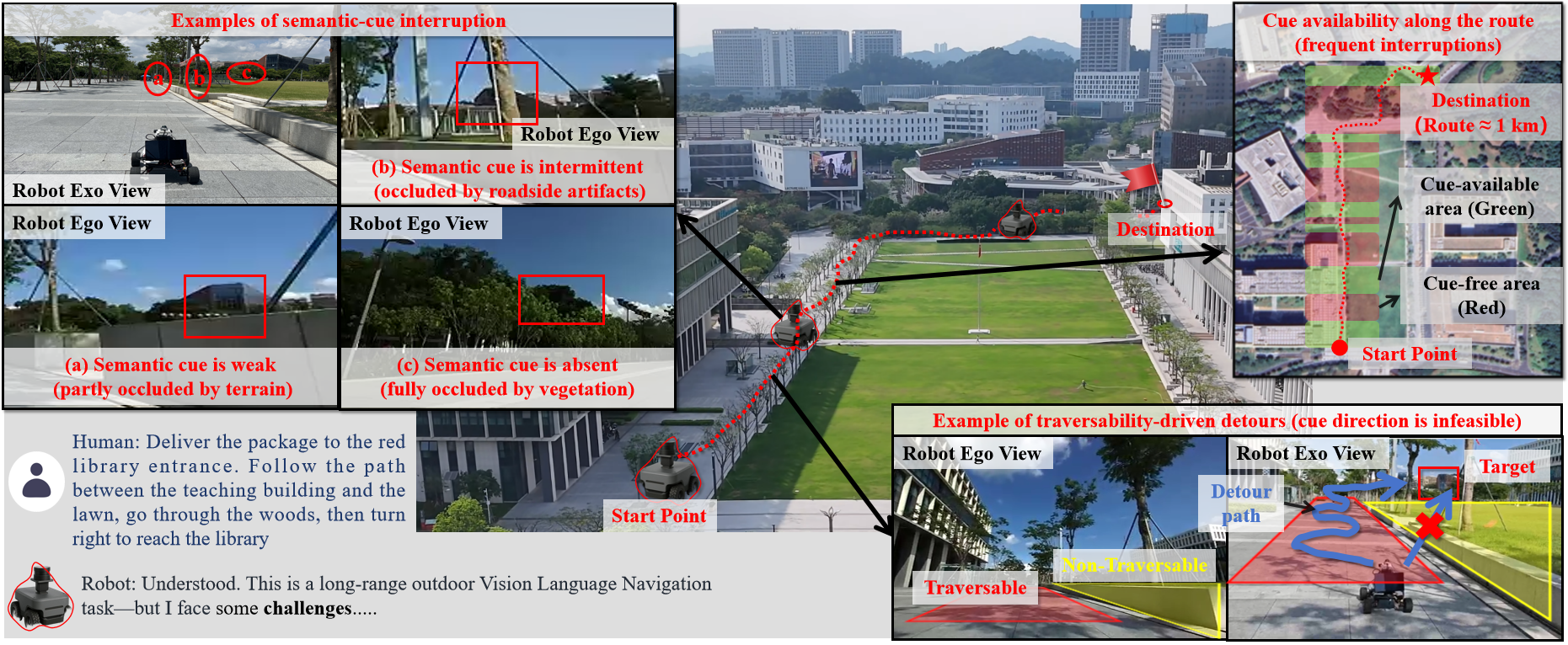}
    \captionof{figure}{\textbf{Surviving Semantic-Cue Interruptions in Outdoor VLN.}
Weak and intermittent semantic evidence from small, distant targets makes navigation cues sparse and repeatedly interrupted, pushing the agent into cue-free phases where it tends to drift or oscillate without stable guidance (left; top-right).
Traversability constraints can further undermine an otherwise correct semantic bearing and induce detours/backtracking, which prolong cue-free navigation and amplify drift (bottom-right).}

    \label{fig:motivation}
    \vspace*{-5mm}
}
\makeatother

\maketitle
\setcounter{figure}{1} 
\thispagestyle{empty}
\pagestyle{empty}

\begin{abstract}
Outdoor vision-language navigation (VLN) in long-range, open-world environments is frequently disrupted by semantic-cue interruptions, where informative goal cues become sparse, occluded, or leave the field of view. Once such cues disappear, agents enter a cue-free phase and often degrade into backtracking, oscillatory headings, or aimless exploration. While memory-based methods attempt to bridge these gaps, they often fail under traversability-driven detours: the remembered cue direction may be infeasible, forcing detours that prolong cue-free phases and gradually render robot-centric cues stale and implicit histories blurred. This makes traversability a stability condition for maintaining goal-directed guidance, rather than merely a local safety concern.

We propose a unified outdoor VLN framework that survives semantic-cue interruptions by maintaining traversability-consistent executable guidance throughout prolonged cue-free phases. Specifically, our method extracts semantic bearings from visibility-gated goal or exploration cues and grounds them into executable headings using a real-time near-field traversability profile, providing goal-consistent feasible guidance beyond reject-only safety filtering. To prevent guidance degradation during detours, we lift intermittent 2D evidence into a world-aligned 3D cue memory with an uncertainty-aware readout mechanism, ensuring guidance remains continuously reachable and stable as the robot moves.

We evaluate the framework on quadrupedal and wheeled platforms over 
600--1000\,m routes. Our method improves simulation success rate by over 
10 percentage points over the strongest baseline and achieves a 
real-world success rate of 40\%, compared to 17.5\% for the strongest 
baseline, with substantially higher robustness during prolonged cue-free 
intervals.
\end{abstract}


\section{Introduction}
\label{sec:intro}
Outdoor vision-language navigation (VLN) is particularly challenging in long-range, open-world scenarios. When informative semantic cues are visible (a cue-available phase), following the instruction is often straightforward; the real watershed is semantic-cue interruption, where such cues can be broken by small distant targets, occlusions, or simply leaving the field of view~\cite{wu2024vision} (a cue-free phase). Once the cue is lost, the agent enters a cue-free phase and often degenerates into backtracking, oscillating headings, or aimless exploration over long horizons.

To survive such interruptions, prior work has explored several common directions. One line re-grounds semantic guidance with VLM priors~\cite{wang2024qwen2,dubey2024llama,team2024gemini,zhang2024vision}, seeking alternative semantic cues to recover goal-directed guidance. This can be effective in indoor settings with dense, strongly correlated semantics, but becomes unreliable outdoors where semantic cues are sparse and weakly correlated. As a result, memory-based strategies have become increasingly common~\cite{wei2025streamvln,zhang2024uni,zheng2025gmm}, maintaining either explicit 2D cost maps/graphs or implicit histories in latent memory spaces to provide guidance after cues disappear.

However, long-range outdoor navigation often breaks such memories in practice due to traversability constraints in real terrain: the remembered cue direction is not necessarily reachable, and the robot must detour or backtrack to remain safe on uneven ground. In many systems~\cite{wang2025genie,elnoor2025vlm}, traversability is enforced as an independent safety filter, which can prevent stepping into unsafe regions but does not provide a goal-consistent alternative when the remembered direction becomes infeasible. 
Such traversability-driven detours increase travel distance and prolong cue-free phases, and can gradually erode both explicit and implicit memories.
In particular, robot-centric direction cues such as keyframe headings or view-dependent focus points can become stale as the robot moves and rotates during long detours, while implicit histories can blur as more irrelevant observations accumulate.
This makes traversability a stability condition for maintaining goal-directed guidance, rather than merely a local safety concern.

Surviving semantic-cue interruptions is therefore central to long-range outdoor VLN. The core question is: how can an agent maintain executable guidance that remains reachable and readable under prolonged cue-free phases?

% Motivated by the above, we propose a unified outdoor VLN framework that survives semantic-cue interruptions by maintaining traversability-consistent executable guidance throughout prolonged cue-free phases. At each step, it extracts a semantic bearing from either an exploration cue or a visibility-gated goal cue, and grounds the bearing into an executable heading using a concurrently estimated near-field traversability profile. To prevent guidance from becoming stale under traversability-driven detours, it further lifts intermittent 2D evidence into a world-aligned 3D cue memory and performs uncertainty-aware cue-free readout, ensuring the guidance remains continuously readable as the robot moves and reachable under traversability constraints.

Motivated by the above, we propose a unified outdoor VLN 
framework that maintains traversability-consistent executable 
guidance throughout prolonged cue-free phases via 
visibility-gated cue extraction, traversability-aware heading 
grounding, and a world-aligned 3D cue memory with 
uncertainty-aware readout.

\section{Related Works}
\label{sec:related}
Outdoor long-range VLN increasingly leverages foundation models to infer semantic guidance from egocentric views, and augments it with mapping, memory, or learned policies to cope with long horizons~\cite{shahvint, sridhar2024nomad,liu2025citywalker}. However, outdoor deployments are frequently dominated by semantic-cue interruptions: goal evidence can be sparse, ambiguous, or temporarily absent due to small distant targets and occlusions, pushing the agent into prolonged cue-free phases where guidance must remain stable while the robot keeps moving.

A common response is to \emph{re-ground} guidance from the current frame by prompting VLM/MLLMs to decide where to look and infer likely directions or regions~\cite{zheng2024towards,sathyamoorthy2024convoi,yuan2025opennav}. Recent methods further couple semantic relevance with geometry by projecting scores onto occupancy maps, frontiers, or exploration candidates~\cite{du2025vl,zeng2025ezreal,alama2025rayfronts}. These strategies can work when semantic evidence is informative, but outdoors the evidence is often weak and weakly correlated across views, so repeatedly re-grounding direction frame by frame can yield oscillatory or drifting bearings during long interruptions. 
% Our approach instead adapts cue extraction to evidence strength: when goal evidence is insufficient it produces instruction-consistent exploration guidance to avoid unstable goal guesses, and when evidence becomes concentrated it extracts reliable goal direction and stores it for guidance during subsequent cue-free phases.

Because per-frame re-grounding can be unreliable, \emph{memory-based designs} have become increasingly common for bridging cue-free phases. Prior work maintains implicit histories in long contexts and hidden states~\cite{wei2025streamvln,zhang2024uni} (or schema-based trajectory memory~\cite{zheng2024towards}), or uses explicit representations such as keyframe target headings and topological structures~\cite{zheng2025gmm,zeng2025ezreal}. These designs can help with short semantic interruptions, but in long-range outdoor runs the robot often detours or backtracks for extended periods, and the stored direction should remain consistent as the robot moves. In this setting, robot-centric cues can become stale over long motions and implicit histories can blur as observations accumulate, leading to unstable guidance in prolonged cue-free phases. 
% We address this by lifting intermittent 2D evidence into a world-aligned 3D cue memory, and using its uncertainty to derive a current direction that stays consistent under robot motion.

Crucially, outdoor execution also requires \emph{terrain feasibility:} the re-grounded or remembered direction may be unsafe or impractical to traverse, forcing detours that can further prolong cue-free phases. Many systems enforce traversability as an external safety layer, e.g., attaching a separate perception module to build a cost field and then filtering actions accordingly~\cite{wang2025genie}. Recent work moves toward tighter integration by estimating traversability with multimodal LLM/VLM prompts to construct risk-aware cost fields~\cite{gummadi2025zest}, or by combining proprioception with VLM predictions to assess future traversability~\cite{elnoor2025vlm}. Yet traversability is still often used in a reject-only manner: it accepts or vetoes a given direction, without providing a goal-consistent alternative when the suggested direction is infeasible. This decoupling can destabilize guidance precisely when cues are missing and detours are unavoidable. 
% In contrast, we explicitly ground semantic bearings into traversability-consistent executable headings at each step, making reachability an intrinsic part of the guidance interface. This treats traversability as a stability condition for maintaining goal-directed guidance, rather than merely a local safety concern.

\section{Method}

\begin{figure*}[!t]
\centering
\includegraphics[width=\linewidth]{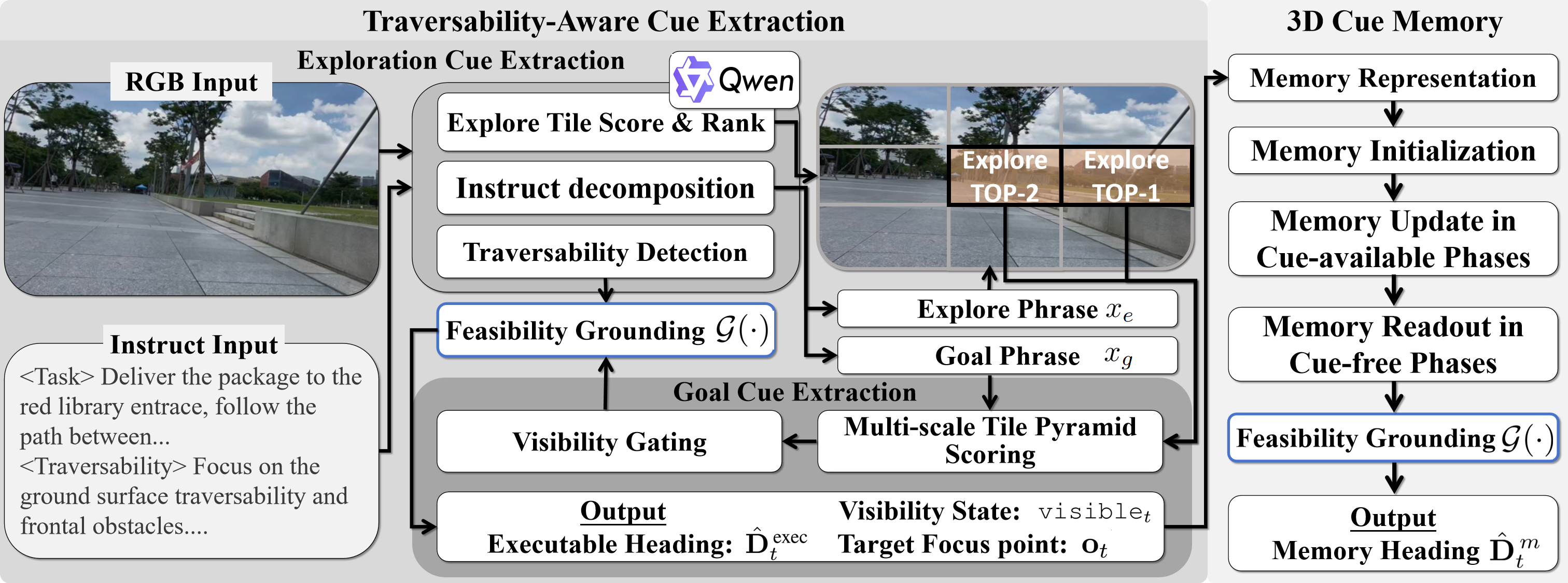}
\caption{\textbf{Method overview.}
Given an RGB frame and an instruction, a Qwen-based VLM decomposes the instruction into an exploration phrase and a goal phrase to drive coarse-to-fine semantic cue extraction. We first perform an exploration-guided coarse extraction to obtain a robust exploration bearing, and then zoom into the most relevant goal-centric region for multi-scale scoring and visibility estimation, yielding a semantic cue direction. In parallel, the same VLM predicts near-field traversability, and a feasibility grounding operator $\mathcal{G}(\cdot)$ maps the cue direction to a feasible executable heading $\hat{\mathbf{D}}^{\,\text{exec}}_t$. The resulting 2D cues further update a particle-based 3D cue memory $\hat{\mathbf{D}}^{\,\text{m}}_t$ that maintains a geometry-consistent belief of the goal; during semantic-cue interruptions, the memory provides world-consistent guidance, bridging cue-available and cue-free phases.}
\vspace{-25pt}
\label{fig:pipeline}
\end{figure*}

\subsection{Traversability-Aware Cue Extraction}
\label{sec:dual_stage_focusing}

Fig.~\ref{fig:pipeline} overviews our per-step data flow. 
To survive semantic-cue interruptions, the agent must maintain executable guidance throughout prolonged cue-free phases. We therefore design traversability-aware cue extraction as a front-end that harvests actionable cues whenever evidence is available and outputs a consistent per-step guidance interface for downstream control and memory update.

At each step $t$, given the current RGB frame $I$ and instruction $x$, we output a semantic bearing $\hat{\mathbf{d}}^{\,\text{cue}}_t$ and its traversability-consistent executable heading $\hat{\mathbf{D}}^{\,\text{exec}}_t$.
The bearing comes from either an exploration cue (when goal evidence is absent) or a visibility-gated goal cue; the heading is obtained by grounding the bearing with a co-inferred estimated near-field traversability profile.
We use lowercase $\hat{\mathbf{d}}^{(\cdot)}_t$ for bearings and uppercase $\hat{\mathbf{D}}^{(\cdot)}_t$ for executable headings.
We parse $x$ into a goal phrase $x_g$ and an exploration phrase $x_e$ 
using a Qwen VLM~\cite{wang2024qwen2} (e.g., for the instruction 
``Deliver the package to the red library entrance, follow the path 
between the teaching building and the lawn, ...'', $x_g\!=\!$ 
``red library entrance'' and $x_e\!=\!$ ``path between teaching 
building and lawn, through the woods''), and use the same VLM 
backend for the per-step queries in this module.

\subsubsection{Exploration cue extraction}
\label{sec:stage1}
When goal evidence is weak, directly predicting a goal 
bearing causes oscillations.
We therefore accumulate an exploration cue to bias motion toward instruction-consistent regions.
We partition $I$ into tiles and rank them by relevance to $x_e$ via $\tau^{K}_{e} = f_{\text{exp}}(I, x_e)$,
where $f_{\text{exp}}(\cdot)$ uses the same Qwen VLM~\cite{wang2024qwen2} to score tiles and returns the top-$K$ set $\tau^{K}_{e}$.
The center ray of the top-ranked tile defines the exploration bearing $\hat{\mathbf{d}}^{\,e}_t$.
Goal cue extraction (Sec.~\ref{sec:stage2}) is then applied only within the image region covered by $\tau^{K}_{e}$.

\subsubsection{Goal cue extraction and visibility gating}
\label{sec:stage2}
This stage outputs a goal bearing only when the evidence is sufficiently concentrated; otherwise it falls back to $\hat{\mathbf{d}}^{\,e}_t$ to avoid hallucination.
Within $\tau^{K}_{e}$, we reuse the multi-scale tile pyramid scoring and peakedness-based visibility test from EzReal~\cite{zeng2025ezreal}, producing a coarse saliency map over tiles and a binary visibility signal $\texttt{visible}_t$. We define $\texttt{visible}_t{=}1$ iff the peakedness ratio $r_t{=}\max(s)/(\overline{s}{+}\epsilon)$ exceeds $\tau_{\text{vis}}{=}1.5$ at the coarse level, i.e., the top-ranked tile is at least $50\%$ above the mean tile score, indicating a concentrated semantic peak rather than a flat distribution.
If $\texttt{visible}_t=1$, we take the center pixel $\mathbf{o}_t$ of the top-ranked coarse tile and back-project it to obtain the goal bearing $\hat{\mathbf{d}}^{\,g}_t$; otherwise we fall back to $\hat{\mathbf{d}}^{\,e}_t$:
\begin{equation}
\hat{\mathbf{d}}^{\,\text{cue}}_t=
\begin{cases}
\hat{\mathbf{d}}^{\,g}_t, & \text{if } \texttt{visible}_t=1,\\
\hat{\mathbf{d}}^{\,e}_t, & \text{if } \texttt{visible}_t=0.
\end{cases}
\end{equation}
We set $\mathbf{o}_t=\varnothing$ when $\texttt{visible}_t=0$.

\begin{figure*}[!t]
\centering
\includegraphics[width=\linewidth]{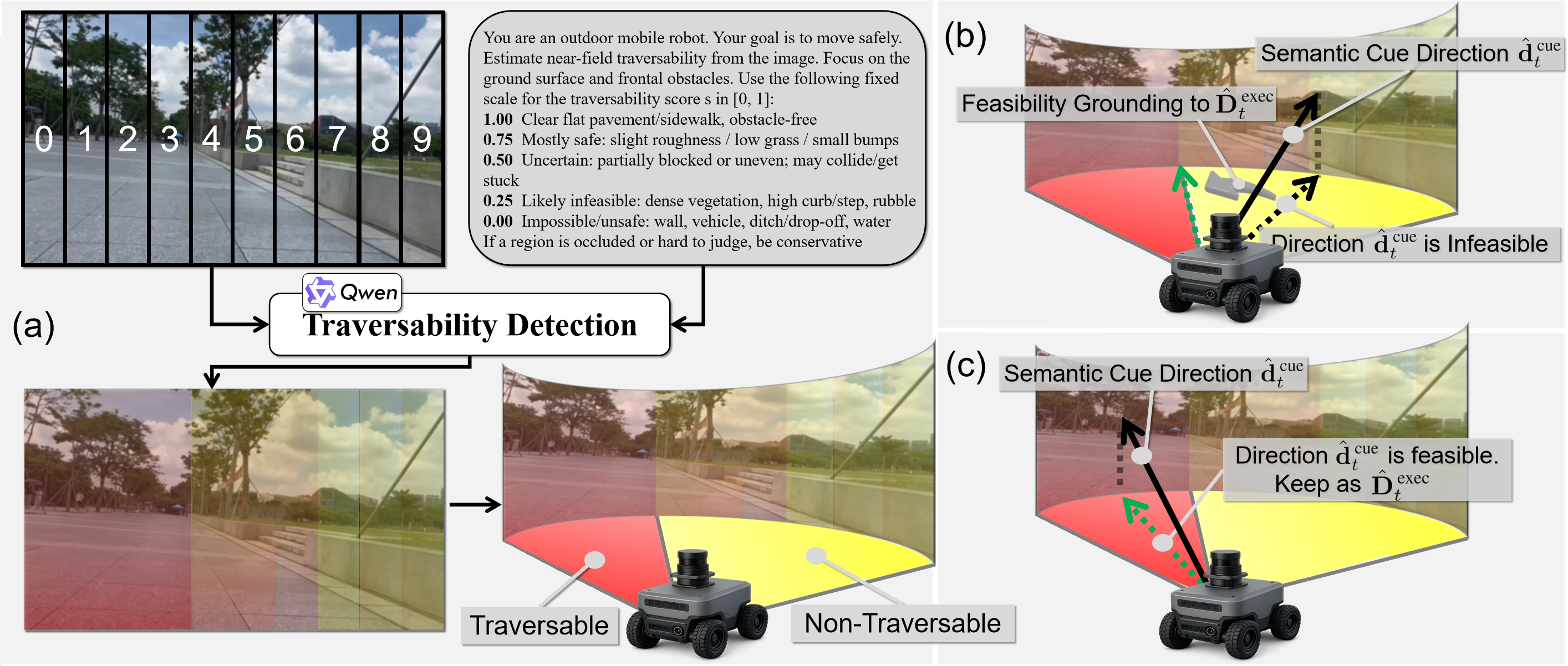}
\caption{\textbf{Traversability-aware heading selection.}
(a) The ego-view is discretized into heading sectors and an instruction-tuned VLM (with a traversability-specific prompt) predicts a traversability score for each sector, visualized as an overlay (red indicates more traversable; yellow indicates less/non-traversable).
(b) If the semantic bearing $\hat{\mathbf{d}}^{\,\text{cue}}_t$ points to a low-traversability sector, traversability grounding $\mathcal{G}(\cdot)$ snaps it to the closest traversable sector and outputs the executable heading $\hat{\mathbf{D}}^{\,\text{exec}}_t$ (green).
(c) If the cue bearing is traversable, it is kept unchanged as $\hat{\mathbf{D}}^{\,\text{exec}}_t$.
This grounding step converts semantic bearings into traversability-consistent executable headings for downstream control.}
\vspace{-25pt}
\label{fig:travel_detect}
\end{figure*}

\subsubsection{Traversability-aware heading grounding}
\label{sec:trav_heading}
A semantic bearing alone is insufficient outdoors: the remembered or newly accumulated bearing may point to terrain that is unsafe or impractical to traverse.
We therefore estimate a traversability profile during cue extraction and ground
$\hat{\mathbf{d}}^{\,\text{cue}}_t$ to a traversability-consistent executable heading:
\begin{equation}
\hat{\mathbf{D}}^{\,\text{exec}}_t=\mathcal{G}\!\left(\hat{\mathbf{d}}^{\,\text{cue}}_t,\{\mu_{t,i},\sigma_{t,i}\}_{i=1}^{N}\right),
\end{equation}
where $\mathcal{G}(\cdot)$ snaps the bearing to the closest traversable sector (Fig.~\ref{fig:travel_detect}).
We discretize the view into $N$ heading sectors and query the same Qwen VLM~\cite{wang2024qwen2} (traversability prompt) for per-sector scores $\{s_{t,i}\}_{i=1}^N$.
We smooth scores with an exponential moving average $\mu_{t,i}=(1-\beta)\mu_{t-1,i}+\beta s_{t,i}$,
and set $\sigma_{t,i}$ as the standard deviation over recent scores.
A sector is traversable if $\mu_{t,i}-\lambda\sigma_{t,i}\ge \tau_{\text{trav}}$,
where $\lambda$ controls risk sensitivity and $\tau_{\text{trav}}$ serves as an interpretable platform knob, enabling the same policy to adapt across robot platforms (Fig.~\ref{fig:traversability}). We use $\lambda{=}1.0$, $\tau_{\text{trav}}{=}0.7$ for the wheeled Scout Mini (which requires firm flat surfaces) and a more permissive $\tau_{\text{trav}}{=}0.5$ for the legged Go1 (which can additionally traverse low grass and small bumps). Both values align with anchors in our traversability prompt (Fig.~\ref{fig:travel_detect}) and are kept fixed across all reported runs.

Finally, we output the per-step interface
\begin{equation}
(\texttt{visible}_t,\hat{\mathbf{D}}^{\,\text{exec}}_t,\mathbf{o}_t)=
\begin{cases}
(1,\hat{\mathbf{D}}^{\,\text{exec}}_t,\mathbf{o}_t),\\
(0,\hat{\mathbf{D}}^{\,\text{exec}}_t,\varnothing).
\end{cases}
\end{equation}
$\hat{\mathbf{D}}^{\,\text{exec}}_t$ can be tracked by any local controller or planner.
It provides the navigation command in cue-available steps (goal-directed when $\texttt{visible}_t=1$) and remains well-defined when goal evidence is weak (exploration-driven when $\texttt{visible}_t=0$).

% This completes per-step cue extraction: we maintain a semantic bearing and ground it to a traversability-consistent executable heading, preventing direction loss during traversability-driven detours. Next, we lift these per-step cues into a persistent world-aligned 3D memory for cue-free readout over long horizons.

\subsection{3D Semantic Cue Memory}
\label{sec:semantic_geometric_memory}

Cue extraction (Sec.~\ref{sec:dual_stage_focusing}) provides per-step executable guidance $\hat{\mathbf{D}}^{\,\text{exec}}_t$ for navigation.
When goal evidence is visible ($\texttt{visible}_t=1$), it also outputs a 2D focus point $\mathbf{o}_t$ that localizes the evidence in the current view.
However, such 2D evidence is still robot-centric and view-dependent: under semantic-cue interruptions, traversability-driven detours and robot motion can make it stale, eroding an executable sense of direction in prolonged cue-free phases.
We therefore lift intermittent focus points into a persistent, world-aligned 3D semantic cue memory. The role of this 3D memory is to provide robust direction-level guidance during cue-free phases, rather than high-precision 3D geometry of the goal.

The memory is updated in cue-available steps ($\texttt{visible}_t=1$).
During cue-free steps ($\texttt{visible}_t=0$), we read out a current semantic bearing from the remembered 3D state and ground it into a traversability-consistent executable heading using the same operator in Sec.~\ref{sec:trav_heading}.

\subsubsection{Memory representation}
We maintain a particle memory $\{(\mathbf{p}_t^m,w_t^m)\}_{m=1}^{M}$ in the world frame, with $\sum_m w_t^m=1$.
Each particle $\mathbf{p}_t^m$ is a hypothesis of the goal's 3D location, accumulated by lifting intermittent 2D focus points across viewpoints.
We read out the 3D estimate by the weighted mean $\hat{\mathbf{p}}_t=\sum_{m=1}^{M} w_t^m \mathbf{p}_t^m$,
and quantify uncertainty by particle dispersion (trace of the weighted covariance, used in Sec.~\ref{sec:mem_readout}).

\subsubsection{Memory initialization}
We bootstrap the memory once goal evidence is visible at two distinct odometry poses.
Each focus point $\mathbf{o}_t$ defines a 3D back-projection ray; we compute $\mathbf{p}_{\text{init}}$ as the midpoint of the shortest segment between the two rays.
Particles ($M{=}200$) are sampled from an isotropic Gaussian 
($\sigma_{\text{init}}{=}1.0$\,m) around $\mathbf{p}_{\text{init}}$ 
with uniform weights. The two bootstrap poses are required to be 
at least $0.5$\,m apart to avoid ill-conditioned triangulation.
We set $\texttt{mem}_t=1$ after initialization (else $\texttt{mem}_t=0$).
If $\texttt{mem}_t=0$, the agent follows the exploration-driven executable heading (i.e., $\hat{\mathbf{D}}^{\,\text{exec}}_t$ when $\texttt{visible}_t=0$) and retries initialization whenever $\texttt{visible}_t=1$.

\subsubsection{Memory update in cue-available phases}
When $\texttt{visible}_t=1$, we update the world-aligned particle memory with a standard particle filter.
Assuming a static goal, we apply a small isotropic diffusion $\mathbf{p}_t^m=\mathbf{p}_{t-1}^m+\boldsymbol{\epsilon}_t^m$, $\boldsymbol{\epsilon}_t^m\!\sim\!\mathcal{N}(\mathbf{0},\sigma_{\text{proc}}^2 \mathbf{I})$
with $\sigma_{\text{proc}}{=}0.1$\,m to maintain particle diversity and absorb noise.
Treating $\mathbf{o}_t$ as a 2D observation, we project each particle to the image to obtain $\mathbf{q}_t^m$ and update weights by a reprojection likelihood:
\begin{equation}
w_t^m \propto w_{t-1}^m \exp\!\Big(-\|\mathbf{q}_t^m-\mathbf{o}_t\|^2/(2\sigma_{\text{obs}}^2)\Big),
\end{equation}
We use $\sigma_{\text{obs}}{=}15$\,px (for a $640{\times}480$ input) 
and trigger systematic resampling when the effective sample size 
drops below $M/2$. Noisy focus points only down-weight inconsistent 
particles rather than collapsing the belief.
We read out $\hat{\mathbf{p}}_t$ by the weighted mean and compute uncertainty $c_t$ from particle dispersion (trace of the weighted covariance).

\subsubsection{Memory readout in cue-free phases}
\label{sec:mem_readout}
When $\texttt{visible}_t=0$, we freeze the last world-aligned memory and continuously convert it into a semantic bearing as the robot moves.
Let $\mathbf{p}^{\text{rob}}_t$ be the odometry position and $\hat{\mathbf{p}}$ the last memory mean. The memory bearing is
\begin{equation}
\hat{\mathbf{d}}^{\,m}_t=\frac{\hat{\mathbf{p}}-\mathbf{p}^{\text{rob}}_t}{\|\hat{\mathbf{p}}-\mathbf{p}^{\text{rob}}_t\|}.
\end{equation}
We use uncertainty $c_t$ to allow larger deviation when the memory is less reliable, by mapping it to a cone half-angle
\begin{equation}
\theta_t=\theta_{\min}+\left(1-\exp(-c_t/\gamma)\right)(\theta_{\max}-\theta_{\min}).
\end{equation}
We then ground $\hat{\mathbf{d}}^{\,m}_t$ to a traversability-consistent executable heading by restricting Sec.~\ref{sec:trav_heading} to the cone:
\begin{equation}
\hat{\mathbf{D}}^{\,m}_{t}=\mathcal{G}_{\theta_t}\!\left(\hat{\mathbf{d}}^{\,m}_t,\{\mu_{t,i},\sigma_{t,i}\}_{i=1}^{N}\right),
\end{equation}
falling back to the exploration-driven $\hat{\mathbf{D}}^{\,exec}_t$ if no traversable heading exists within the cone.

This cue-free readout keeps guidance reachable under traversability constraints while remaining readable under robot motion.

\section{Experimental Results}
\label{sec:exp}

\begin{table*}[t]
    \centering
    \caption{Overall performance and cue-free survival statistics in simulation scenes. SR, Fail@CF and CFSR@D are reported in \%; SPL is unitless. All values are mean$\pm$std over 3 random seeds 
on 100 episodes per scene.}
    \label{tab:sim_main_cuefree}
    \footnotesize
    \setlength{\tabcolsep}{2.5pt}
    \renewcommand{\arraystretch}{1.06}
    \begin{tabular*}{\textwidth}{@{\extracolsep{\fill}}lcccccccccccc}
        \toprule
        \multirow{4}{*}{Method}
        & \multicolumn{6}{c}{Wild}
        & \multicolumn{6}{c}{Suburban} \\
        \cmidrule(lr){2-7} \cmidrule(lr){8-13}
        & \multirow{2}{*}{SR $\uparrow$}
        & \multirow{2}{*}{SPL $\uparrow$}
        & \multirow{2}{*}{Fail@CF $\downarrow$}
        & \multicolumn{3}{c}{CFSR@D $\uparrow$}
        & \multirow{2}{*}{SR $\uparrow$}
        & \multirow{2}{*}{SPL $\uparrow$}
        & \multirow{2}{*}{Fail@CF $\downarrow$}
        & \multicolumn{3}{c}{CFSR@D $\uparrow$} \\
        \cmidrule(lr){5-7} \cmidrule(lr){11-13}
        &  &  &  & 10m & 25m & 50m
        &  &  &  & 10m & 25m & 50m \\
        \midrule
        NoMaD~\cite{sridhar2024nomad}
            & 6.3$\pm$2.5  & 0.14$\pm$0.02 & 86.0$\pm$2.0 & 26.0$\pm$3.0 & 3.7$\pm$1.5  & 1.3$\pm$1.5
            & 14.3$\pm$3.5 & 0.18$\pm$0.03 & 78.0$\pm$3.0 & 47.0$\pm$3.0 & 12.3$\pm$2.5 & 8.0$\pm$3.0 \\
        NaviLLM~\cite{zheng2024towards}
            & 12.3$\pm$3.5 & 0.25$\pm$0.03 & 84.0$\pm$4.0 & 34.0$\pm$4.0 & 12.3$\pm$2.5 & 6.7$\pm$2.5
            & 22.3$\pm$4.5 & 0.24$\pm$0.04 & 71.0$\pm$4.0 & 50.3$\pm$4.5 & 20.0$\pm$4.0 & 20.3$\pm$4.5 \\
        StreamVLN~\cite{wei2025streamvln}
            & 14.7$\pm$4.0 & 0.23$\pm$0.04 & 82.0$\pm$4.0 & 39.3$\pm$4.5 & 14.0$\pm$3.0 & 11.7$\pm$4.0
            & 24.7$\pm$5.0 & 0.25$\pm$0.03 & 64.0$\pm$3.0 & 48.0$\pm$4.0 & 22.0$\pm$4.0 & 19.3$\pm$3.5 \\
        GPT-4o-GeNie~\cite{wang2025genie}
            & 22.0$\pm$5.0 & 0.31$\pm$0.04 & 72.0$\pm$4.0 & 52.0$\pm$5.0 & 25.0$\pm$4.0 & 20.3$\pm$4.5
            & 27.3$\pm$5.5 & 0.35$\pm$0.03 & 58.0$\pm$4.0 & 60.0$\pm$5.0 & 30.0$\pm$4.0 & 28.0$\pm$4.0 \\
        EzReal~\cite{zeng2025ezreal}
            & 29.0$\pm$3.0 & 0.18$\pm$0.03 & 60.0$\pm$3.0 & 50.0$\pm$3.0 & 37.3$\pm$2.5 & 27.3$\pm$3.5
            & 33.0$\pm$4.0 & 0.34$\pm$0.02 & 50.3$\pm$3.5 & 65.0$\pm$3.0 & 43.0$\pm$3.0 & 35.7$\pm$2.5 \\
        \textbf{Ours}
            & \textbf{41.0$\pm$4.0} & \textbf{0.35$\pm$0.03} & \textbf{41.0$\pm$4.0} & \textbf{84.7$\pm$4.5} & \textbf{64.0$\pm$4.0} & \textbf{52.0$\pm$5.0}
            & \textbf{50.0$\pm$5.0} & \textbf{0.40$\pm$0.03} & \textbf{27.0$\pm$3.0} & \textbf{92.0$\pm$4.0} & \textbf{75.0$\pm$4.0} & \textbf{64.0$\pm$4.0} \\
        \bottomrule
    \end{tabular*}
    \vspace{-10pt}
\end{table*}

\begin{table}[t]
    \centering
    \caption{Comparison of traversability estimation backends in the Wild scene.
    We report IoU against simulator GT, along with deployment-relevant latency and availability.}
    \label{tab:trav_vlm_iou}
    \footnotesize
    \renewcommand{\arraystretch}{1.15}
    \setlength{\tabcolsep}{6pt}
    \begin{tabular*}{\columnwidth}{@{\extracolsep{\fill}}lccc}
        \toprule
        Method & IoU $\uparrow$ & Latency (s) $\downarrow$ & Deployment \\
        \midrule
        Qwen 3 VL~\cite{bai2025qwen3vltechnicalreport}   & 0.74 & $\sim$2.0 & Local \\
        GPT-4o~\cite{openai2024gpt4o}              & 0.85 & $>$5.0    & Online \\
        Gemini 2.5~\cite{comanici2025gemini}               & 0.82 & $>$5.0    & Online \\
        \bottomrule
    \end{tabular*}
    \vspace{-20pt}
\end{table}

\begin{table}[t]
    \centering
    \caption{Overall performance in real-world deployment across two 
    campus routes ($20$ episodes per route, $40$ in total) on the 
    Scout Mini platform.}
    \label{tab:real_main}
    \scriptsize
    \renewcommand{\arraystretch}{1.05}
    \setlength{\tabcolsep}{2.5pt}
    \begin{tabular*}{\columnwidth}{@{\extracolsep{\fill}}lcccccc}
        \toprule
        \multirow{2}{*}{Method}
        & \multicolumn{2}{c}{Route~1}
        & \multicolumn{2}{c}{Route~2}
        & \multicolumn{2}{c}{Combined} \\
        \cmidrule(lr){2-3} \cmidrule(lr){4-5} \cmidrule(lr){6-7}
        & SR$\uparrow$ & SPL$\uparrow$
        & SR$\uparrow$ & SPL$\uparrow$
        & SR$\uparrow$ & SPL$\uparrow$ \\
        \midrule
        NoMaD~\cite{sridhar2024nomad}     & 0.0  & 0.00 & 5.0  & 0.10 & 2.5  & 0.05 \\
        NaviLLM~\cite{zheng2024towards}   & 5.0  & 0.12 & 10.0 & 0.16 & 7.5  & 0.14 \\
        StreamVLN~\cite{wei2025streamvln} & 0.0  & 0.00 & 10.0 & 0.14 & 5.0  & 0.07 \\
        GPT-4o-GeNie~\cite{wang2025genie} & 15.0 & 0.23 & 20.0 & 0.28 & 17.5 & 0.26 \\
        EzReal~\cite{zeng2025ezreal}      & 10.0 & 0.18 & 20.0 & 0.26 & 15.0 & 0.22 \\
        \textbf{Ours}                     & \textbf{35.0} & \textbf{0.28}
                                          & \textbf{45.0} & \textbf{0.36}
                                          & \textbf{40.0} & \textbf{0.32} \\
        \bottomrule
    \end{tabular*}
    \vspace{-25pt}
\end{table}

% 真实实验可视化，这里要把细节的信息标注加上去
\begin{figure*}[!t]
\centering
\includegraphics[width=\linewidth]{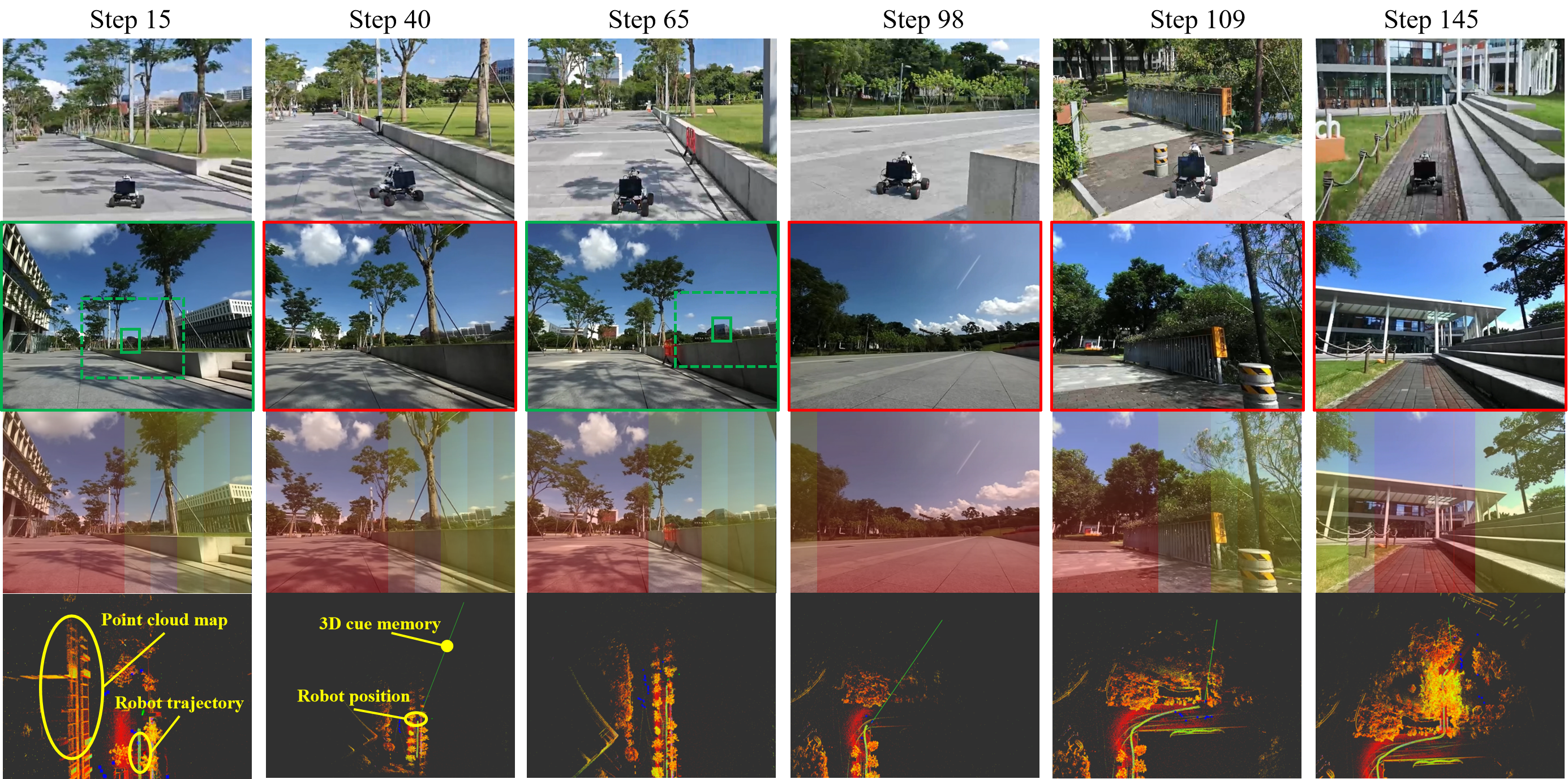}
\caption{\textbf{Visualization of real-world deployment.} Columns show selected timesteps.
\textbf{Top row:} third-person views of the robot.
\textbf{Second row:} ego-centric RGB views. A green border indicates cue-available and a red border indicates cue-free (semantic-cue interruption).
Cue extraction is visualized with a dashed box for exploration cue and a solid box for goal cue.
\textbf{Third row:} predicted traversability overlaid on the RGB view, where warmer colors denote more traversable regions.
\textbf{Bottom row:} RViz visualization of the 3D cue memory; the green ray points to the current 3D belief of the goal in the world frame.
At steps 40, 98, 109, and 145 the semantic cue is interrupted; the policy follows the 3D memory-based bearing (green ray updated as the robot moves) and selects traversability-consistent executable headings until the cue is re-acquired.
The point-cloud map and blue markers in RViz are provided by the onboard mapping/planning stack for visualization only and are not part of our method.}

\vspace{-15pt}
\label{fig:realexp_vis}
\end{figure*}

\begin{figure}[!t]
\centering
\includegraphics[width=\linewidth]{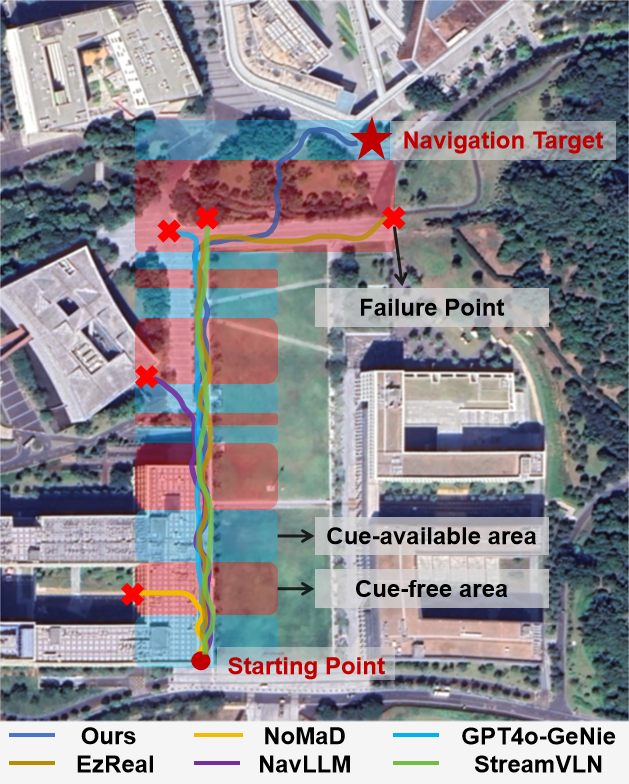}
\caption{\textbf{Real-world trajectories with semantic-cue interruptions.}
Blue/red segments indicate cue-available/cue-free phases along the route.
Curves show trajectories of our method and baselines; crosses mark representative failure locations.
Baselines often terminate in cue-free segments after losing goal-consistent guidance, while our method maintains progress and reaches the target.
Prompt in this route:\textit{``Deliver the package to the red library 
entrance. Follow the path between the teaching building and the lawn, 
go through the woods, then turn right to reach the library.''} 
}

\vspace{-25pt}
\label{fig:realexp_gps_vis}
\end{figure}

\subsection{Simulation Experiments}
\label{sec:sim_exp}
We evaluate the proposed method in a controlled Unreal Engine simulation designed to reproduce semantic-cue interruptions in long-range outdoor VLN.

\subsubsection{Environment and Task Setup}

\begin{figure}[!t]
\centering
\includegraphics[width=\linewidth]{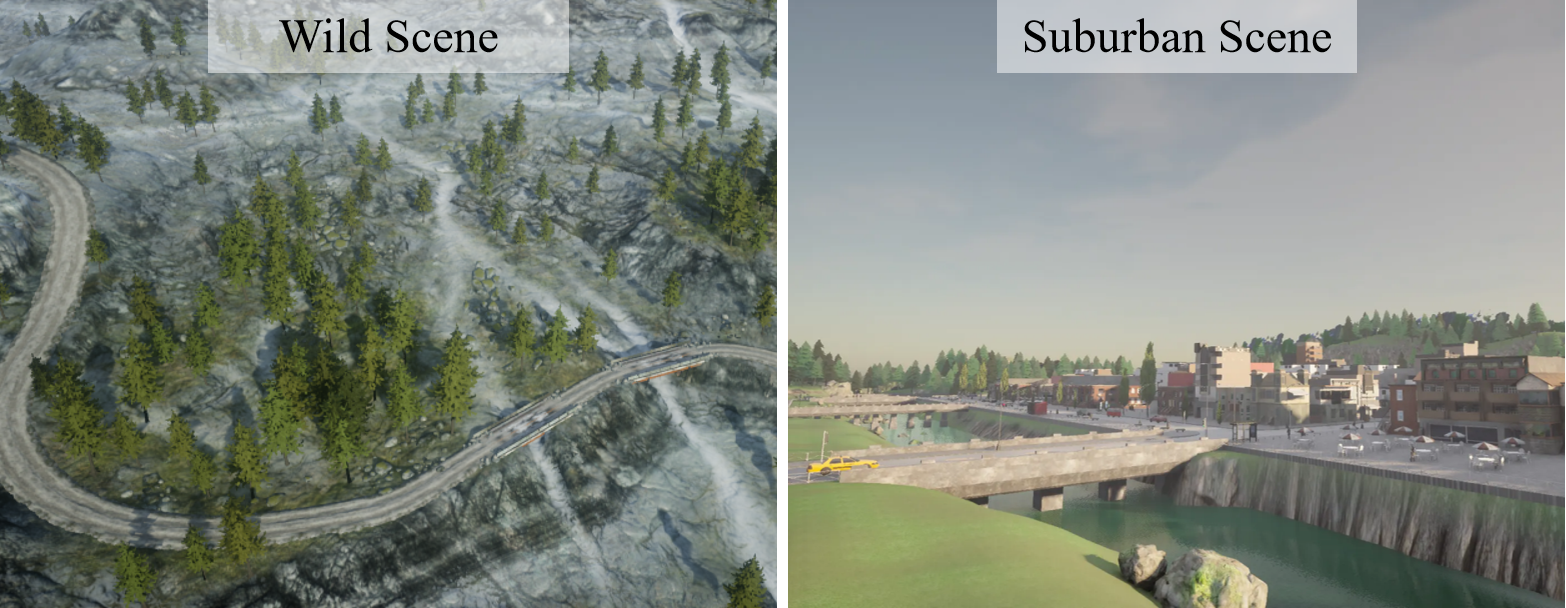}
\caption{Simulation experiment environment (scene view)}
\vspace{-25pt}
\label{fig:sim_env}
\end{figure}

We construct two outdoor scenes (Fig.~\ref{fig:sim_env}): \textbf{Wild}, where sparse distant targets and frequent occlusions are coupled with large non-traversable regions such as rough terrain and blocked areas, forcing agents to take detours and backtrack, which in turn prolongs cue-free phases, and \textbf{Suburban}, characterized by structured roads with continuous cues, yet featuring local traversability challenges like curbs and narrow passages.
Targets are placed hundreds of meters away (typically $600$--$1000\,\mathrm{m}$ straight-line distance), so agents often navigate through extended cue-free stretches where the semantic cue is weak, occluded, or out of view.
All tasks use single-goal, long-range instructions.

\textbf{Evaluation Protocol.} 
To ensure controlled and reproducible comparisons, each simulation 
scene is evaluated on 100 episodes. For each episode, a starting 
pose is randomly sampled from traversable regions and the agent is 
tasked with navigating to one of 3 distinct points-of-interest 
designated as goals in the scene, ensuring straight-line distances 
of 600--1000\,m and a mixture of cue-available and cue-free starts. 
The exact same sequence of (start, goal) configurations is replayed 
across all baselines and our method, so that the navigation policy 
itself is the only experimental variable. To characterize stochastic 
variability arising from particle filtering and VLM sampling, each 
method is evaluated under 3 random seeds; we report mean$\pm$std in 
Table~\ref{tab:sim_main_cuefree}.

\subsubsection{Baselines and Metrics}
All VLM modules in our method use Qwen3-VL-4B~\cite{bai2025qwen3vltechnicalreport}.
We compare with NoMaD~\cite{sridhar2024nomad} (end-to-end visual policy), NaviLLM~\cite{zheng2024towards} (frame-wise VLM-based direction prediction), StreamVLN~\cite{wei2025streamvln} (sequence-based VLN with implicit memory), EzReal~\cite{zeng2025ezreal} (2D small-object sensing and direction memory), and GPT-4o-GeNie. Since GeNie's official release contains only the traversability-aware planner, we pair it with GPT-4o as the perception front-end: GPT-4o localizes the target in each RGB frame, the predicted pixel is back-projected to a goal bearing, and the bearing is passed unchanged to GeNie's planner. All baselines use their official open-source configurations under default settings.

We report \textbf{Success Rate (SR)}, where an episode is successful if the final distance to the target is below $3\,\mathrm{m}$ and the target is visible without full occlusion, and \textbf{Success weighted by Path Length (SPL)}.

To directly evaluate robustness to semantic-cue interruptions, we additionally report cue-free failure attribution and survival statistics. Using simulator ground-truth target visibility, we mark each step as cue-available or cue-free. For failed episodes, \textbf{Fail@CF} measures the fraction of failures whose termination occurs during cue-free phases. 

We further quantify how long an agent remains goal-directed after entering a cue-free period by reporting the \textbf{cue-Free Survival Ratio (CFSR)} at multiple distances (CFSR@10/25/50\,m).
Let $k$ denote the last time when the target is visible before termination (or $k=0$ if the target is not observed).
Let $r_t=\|\mathbf{p}_t-\mathbf{g}\|$ be the Euclidean distance from the robot position $\mathbf{p}_t$ to the goal position $\mathbf{g}$, and let $\ell_{\bar{t}}$ be the accumulated traveled path length.
We detect the first time the agent stops making progress toward the goal using a fixed window $W$:
\begin{equation}
\bar{t}=\min\{t\ge k+W \mid r_t \ge r_{t-W}\},
\end{equation}
If the condition is never met before termination, we set $\bar{t}=T$, where $T$ is the termination time index. We then define the effective cue-free survival distance as $L_{\text{survive}} = \ell_{\bar{t}} - \ell_{k}$.
For a distance threshold $D$, we compute
\begin{equation}
\text{CFSR@}D=\frac{1}{|\mathcal{F}|}\sum_{e\in\mathcal{F}}\mathbb{I}\big[L_{\text{survive}}(e)\ge D\big],
\end{equation}
where $\mathcal{F}$ is the set of failed episodes. 

% 模拟环境图

\subsubsection{Implementation Details}
\label{sec:hp_sensitivity}
Hyperparameters are listed in Table~\ref{tab:hyperparams} and kept 
fixed across all reported runs. Three complementary sensitivity 
views are reported: platform sensitivity via the 
$\tau_{\text{trav}}$ switch between legged ($0.5$) and wheeled 
($0.7$) robots (Fig.~\ref{fig:traversability}); backend sensitivity 
via VLM substitution for traversability estimation 
(Table~\ref{tab:trav_vlm_iou}); and structural sensitivity via 
per-module ablation (Table~\ref{tab:ablation}).

\subsubsection{Results}
Table~\ref{tab:sim_main_cuefree} shows that our method achieves the best SR/SPL in both scenes and substantially stronger cue-free survival.

Cue-free metrics highlight three recurring failure patterns as interruptions grow longer. 
(i) Frame-wise re-grounding (e.g., NaviLLM~\cite{zheng2024towards}) is easily distracted by weak semantics once the cue disappears, leading to heading flips and drifting exploration. 
(ii) Memory-based designs (e.g., StreamVLN~\cite{wei2025streamvln} and EzReal~\cite{zeng2025ezreal}) are more stable at short ranges but degrade under long detours: robot-centric cues become stale with motion and implicit histories blur as irrelevant observations accumulate, often causing overshoot without recovery. 
(iii) Feasibility used in a reject-only manner (e.g., GPT-4o-GeNie~\cite{wang2025genie}) can further destabilize cue-free behavior: after vetoing an infeasible suggestion, the policy falls back to safe-but-not-goal-consistent local detours, which prolong cue-free exposure and dilute guidance.

Our design directly targets each of these patterns 
via visibility-gated extraction, traversability grounding, 
and world-aligned 3D memory readout.
% Our design directly targets these patterns. 
% Visibility-gated goal extraction with an exploration fallback suppresses distractor locking and heading flips; traversability grounding provides a goal-consistent executable alternative rather than a veto; and the world-aligned 3D belief supports motion-consistent readout, preventing stale robot-centric headings and preserving net progress over extended cue-free distances.

\subsubsection{Capability analysis of traversability estimation}
\label{sec:trav_vlm_compare}
\begin{figure}[!t]
\centering
\includegraphics[width=\linewidth]{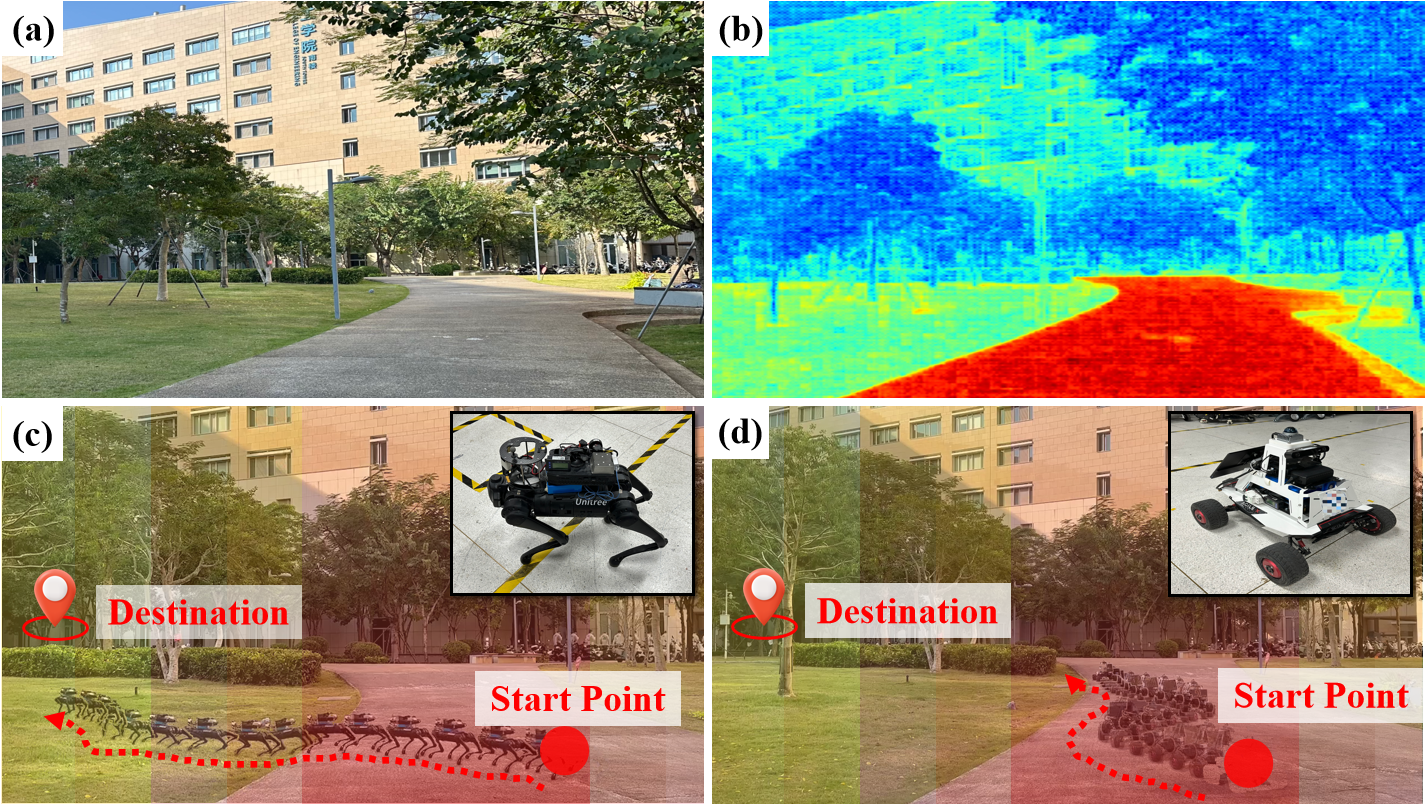}
\caption{\textbf{Traversability-aware heading selection and platform adaptation.} (a) Campus scene. (b) Human-annotated traversability map (warmer: more traversable). (c,d) Predicted traversability overlays and resulting trajectories to the same destination: with $\tau_{\text{trav}}{=}0.5$, the legged Go1 cuts across grass for a shorter route (c), while $\tau_{\text{trav}}{=}0.7$ keeps the wheeled Scout Mini on paved paths (d).
Prompt in this route:\textit{``Find the wooden pavilion across the open 
lawn. Walk along the pedestrian path, cross the grass field while 
staying near the trees, and continue toward the pavilion ahead.''} 
}

\vspace{-25pt}
\label{fig:traversability}
\end{figure}

We analyze the sensitivity of our traversability estimation module (Sec.~\ref{sec:trav_heading}) to the choice of VLM backend under deployment-relevant constraints.
In the Wild simulation scene, we replace the Qwen VLM~\cite{bai2025qwen3vltechnicalreport} with alternative backends (GPT-4o~\cite{openai2024gpt4o} and Gemini~\cite{team2024gemini}), while keeping the rest of the pipeline unchanged.
For evaluation, we render the sector-wise outputs into binary traversability masks and compare them against simulator-provided ground truth using intersection over union (IoU).
% As summarized in Table~\ref{tab:trav_vlm_iou}, different backends exhibit bounded IoU variations while differing substantially in inference latency and deployability, indicating that our feasibility grounding interface is not tied to a specific model.
% In practice, we adopt Qwen due to its low-latency, fully local deployment, which provides sufficiently reliable feasibility signals for grounding semantic bearings into traversability-consistent executable headings.
% Fig.~\ref{fig:traversability} further provides qualitative evidence from real-world deployment that the predicted traversability yields sensible heading choices in practice.
Table~\ref{tab:trav_vlm_iou} shows bounded IoU variation across backends, confirming the grounding interface is model-agnostic. 
We adopt Qwen for its local, low-latency deployment; 
Fig.~\ref{fig:traversability} provides qualitative 
validation in real-world conditions.

% 可通行性真实实验图

\subsection{Real-World Experiments}
\label{sec:real_exp}
After the simulation studies, we evaluate the proposed method in a real outdoor environment to assess robustness to semantic-cue interruptions.

\subsubsection{Environment and Deployment}
Experiments are conducted on two distinct campus routes spanning 
$600$--$1000\,\mathrm{m}$ from start to goal. \textbf{Route~1:} 
a paved campus route (Fig.~\ref{fig:motivation}) with roads, curbs, 
vegetation, parked vehicles, and buildings. \textbf{Route~2:} a 
lawn-area route with open grass, scattered trees, and pedestrian 
paths, also depicted qualitatively in Fig.~\ref{fig:traversability}. 

Each route corresponds to a single goal location with a distinct 
natural-language mission prompt (given in Fig.~\ref{fig:realexp_gps_vis} and Fig.~\ref{fig:traversability}), with starting poses distributed across the surrounding area. 
Both routes include multiple long interruption segments where the cue 
is occluded or out of view, yielding extended cue-free phases.

We conduct 20 episodes per route from distinct starting poses 
pre-defined across the surrounding area, with the same set of 
starting poses used across all baselines and our method (40 
episodes in total) on the Scout Mini wheeled platform, equipped with a 
forward-facing stereo camera and LiDAR (extrinsics are pre-calibrated by~\cite{zeng2025yoco}). Inference runs on an external 
PC with an RTX 3090 GPU. The Unitree Go1 quadruped is 
additionally used for qualitative platform-adaptation demonstration in 
Fig.~\ref{fig:traversability}, rather than as a separate quantitative 
platform.

\subsubsection{Tasks, Baselines, and Metrics}
We use the same long-range single-goal VLN instructions as in simulation, with mixed starting poses that include cue-available and cue-free starts.
Baselines follow Sec.~\ref{sec:sim_exp}.
We report SR and SPL; any operator takeover terminates the episode and is counted as failure.
Takeovers are triggered only to prevent entering clearly non-traversable areas, enforcing a consistent feasibility constraint across methods.

\subsubsection{Results}
Table~\ref{tab:real_main} reports the real-world results. Our method 
achieves the highest SR and SPL on both routes, with a combined SR 
of $40.0\%$ over the 40 episodes.

Fig.~\ref{fig:realexp_gps_vis} highlights a consistent cue-free failure pattern: most baselines terminate inside cue-free segments after losing a stable, reachable direction. 
% Re-grounding-heavy methods (e.g., NaviLLM~\cite{zheng2024towards} and GPT-4o-GeNie~\cite{wang2025genie}) are easily distracted once the cue disappears, drifting into wrong corridors and failing to re-align with the goal. 
% Memory-based methods (e.g., StreamVLN~\cite{wei2025streamvln} and EzReal~\cite{zeng2025ezreal}) often degrade into robot-centric ``frozen'' guidance that does not stay consistent under robot motion, leading to overshoot or irreversible drift in prolonged interruptions.

% In contrast, our trajectory remains goal-consistent through long cue-free regions because guidance is continuously read out from a world-aligned 3D belief and grounded into traversability-consistent executable headings, enabling purposeful detours instead of accumulating drift.
The failure patterns mirror those in simulation 
(Sec.~\ref{sec:sim_exp}): baselines lose goal-consistent 
guidance in cue-free phases, while our method maintains progress 
via world-aligned memory readout.

\subsubsection{Runtime Analysis}
The system runs as a three-layer stack with increasing rates: 
(i) a VLM perception layer at $0.3$–$0.5$\,Hz on an RTX 3090 
($\sim 2.8$\,s per update), dominated by Qwen inference; 
(ii) a guidance-integration layer at $\sim 1$\,Hz that combines 
the latest VLM outputs with the 3D memory readout and the 
traversability profile to produce $\hat{\mathbf{D}}^{\,\text{exec}}_t$; 
and (iii) the platform's native low-level controller (around 
$50$\,Hz on Scout Mini), which tracks $\hat{\mathbf{D}}^{\,\text{exec}}_t$ 
and keeps motion commands smooth. As a result, the reactive 
horizon to fast near-field changes is bounded by the VLM 
refresh rate, which we discuss as failure mode~(4) in Sec.~V. 
Standard techniques such as quantization and pipeline 
parallelism can reduce VLM latency without changing this 
three-layer interface.

\begin{table}[t]
    \centering
    \caption{Ablation results in the Wild scene. SR, Fail@CF and CFSR@50 are reported in \%; SPL is unitless.}
    \label{tab:ablation}
    \scriptsize
    \setlength{\tabcolsep}{2pt}
    \renewcommand{\arraystretch}{1.05}
    \begin{tabular*}{\columnwidth}{@{\extracolsep{\fill}}lcccc}
        \toprule
        Variant & SR $\uparrow$ & SPL $\uparrow$ & Fail@CF $\downarrow$ & CFSR@50 $\uparrow$ \\
        \midrule
        w/o Cue Extraction          & 32.0$\pm$4.0 & 0.27$\pm$0.03 & 64.0$\pm$4.0 & 24.0$\pm$4.0 \\
        w/o 3D Cue Memory           & 30.0$\pm$4.0 & 0.26$\pm$0.03 & 72.0$\pm$4.0 & 16.0$\pm$4.0 \\
        w/o Trav. Grounding         & 36.0$\pm$4.0 & 0.22$\pm$0.03 & 54.0$\pm$4.0 & 30.0$\pm$4.0 \\
        \midrule
        Full system                 & 41.0$\pm$4.0 & 0.35$\pm$0.03 & 41.0$\pm$4.0 & 52.0$\pm$5.0 \\
        \bottomrule
    \end{tabular*}
    \vspace{-20pt}
\end{table}

\subsection{Ablation Studies}
\label{sec:ablation}
We conduct ablations in the harder Wild scene to quantify how each core module supports cue-free survival under semantic-cue interruptions. The setup follows Sec.~\ref{sec:sim_exp}. We report SR/SPL and the interruption-specific metrics Fail@CF and CFSR@50.

We evaluate three variants aligned with our method design.
\textbf{w/o Cue Extraction} removes Sec.~\ref{sec:dual_stage_focusing} (no exploration fallback);
\textbf{w/o 3D Cue Memory} replaces 
the world-aligned particle memory (Sec.~\ref{sec:semantic_geometric_memory}) 
with robot-centric 2D memory;
\textbf{w/o Traversability Grounding} 
removes $\mathcal{G}(\cdot)$ (Sec.~\ref{sec:trav_heading}) and uses semantic bearings directly as headings.

Table~\ref{tab:ablation} shows that the three components fail in complementary ways, each breaking a key link of the guidance loop.
Without 3D cue memory, guidance becomes robot-centric and stale under motion (Fail@CF $72.0$, CFSR@50 $16.0$);
without traversability grounding, semantic bearings often become infeasible and inflate detours (SPL $0.22$);
without cue extraction, the policy relies on unstable goal guesses under weak evidence (SR $32.0$, CFSR@50 $24.0$).

%% 参数表
% \begin{table}[t]
%     \centering
%     \caption{Hyperparameter settings, fixed across all reported runs.}
%     \label{tab:hyperparams}
%     \footnotesize
%     \renewcommand{\arraystretch}{1.0}
%     \setlength{\tabcolsep}{6pt}
%     \begin{tabular*}{\columnwidth}{@{\extracolsep{\fill}}ll}
%         \toprule
%         Parameter & Value \\
%         \midrule
%         $K$ / $N$ / $\beta$ / $\lambda$ & $2$ / $10$ / $0.5$ / $1.0$ \\
%         $\tau_{\text{trav}}$ (Scout / Go1) & $0.7$ / $0.5$ \\
%         $\tau_{\text{vis}}$ & $1.5$ \\
%         $M$ / ESS thr. & $200$ / $0.5M$ \\
%         $\sigma_{\text{proc}}$ / $\sigma_{\text{obs}}$ / $\sigma_{\text{init}}$ & $0.1$\,m / $15$\,px / $1.0$\,m \\
%         Min. baseline & $0.5$\,m \\
%         $\theta_{\min}$ / $\theta_{\max}$ / $\gamma$ & $15°$ / $60°$ / $1.0$\,m$^2$ \\
%         $W$ & $5$ steps \\
%         \bottomrule
%     \end{tabular*}
%     \vspace{-10pt}
% \end{table}
\begin{table}[t]
    \centering
    \caption{Hyperparameter settings, fixed across all reported runs.}
    \label{tab:hyperparams}
    \footnotesize
    \renewcommand{\arraystretch}{0.85}
    \setlength{\tabcolsep}{4pt}
    \begin{tabular*}{\columnwidth}{@{\extracolsep{\fill}}ll}
        \toprule
        Parameter & Value \\
        \midrule
        $K$ / $N$ / $\beta$ / $\lambda$ & $2$ / $10$ / $0.5$ / $1.0$ \\
        $\tau_{\text{trav}}$ (Scout/Go1) / $\tau_{\text{vis}}$ & $0.7$ / $0.5$ / $1.5$ \\
        $M$ / ESS thr. / Min. baseline & $200$ / $0.5M$ / $0.5$\,m \\
        $\sigma_{\text{proc}}$ / $\sigma_{\text{obs}}$ / $\sigma_{\text{init}}$ & $0.1$\,m / $15$\,px / $1.0$\,m \\
        $\theta_{\min}$ / $\theta_{\max}$ / $\gamma$ / $W$ & $15^\circ$ / $60^\circ$ / $1.0$\,m$^2$ / $5$ steps \\
        \bottomrule
    \end{tabular*}
    \vspace{-20pt}
\end{table}

\section{Limitations and Conclusions}
\label{sec:con}
We presented TARIC, a unified outdoor VLN framework that treats 
traversability as a stability condition rather than a safety 
filter, and maintains goal-directed guidance through prolonged 
cue-free phases via traversability-aware cue extraction and a 
world-aligned 3D cue memory. Quantitative evaluations in simulation and on a wheeled platform, together with qualitative quadruped adaptation, confirm substantial gains in success rate and cue-free robustness over strong baselines.

\textbf{Failure modes.}
Despite the gains, the system exhibits several recurring failure 
patterns that we observed across simulation and real-world runs:

\textit{(1) Cue detection and gating errors.}
At extreme range or under harsh lighting, the VLM occasionally 
misses small distant targets entirely, leaving the robot in 
exploration mode for far longer than necessary. The visibility 
gate also produces occasional false positives on visually 
cluttered scenes — for example, distant signage or vegetation 
that briefly forms a strong saliency peak similar to the goal — 
which can pull the memory toward a spurious 3D location before 
later observations down-weight it. 

\textit{(2) Memory drift under sustained cue-free intervals.}
The drift-cancellation in our 3D cue memory relies on 
cue-available updates occurring within a reasonable horizon. 
On real-world routes with long featureless detours — typically 
50\,m or more without any goal observation, often along visually 
repetitive corridors where the LiDAR-inertial stack accumulates 
error without loop closure — the world-aligned mean 
$\hat{\mathbf{p}}$ drifts away from the current robot frame. 
At simple T-junctions this rarely changes the outcome, but at 
multi-way junctions a few degrees of bearing deviation is 
enough to commit the agent to the wrong corridor; once 
committed, recovery cost is high and the episode usually fails.

\textit{(3) Suboptimal traversability grounding.}
The per-frame VLM traversability estimate is reliable on 
clearly traversable or clearly blocked terrain, but degrades on 
ambiguous regions: thin grass partly covering pavement, shaded 
or wet surfaces, low curbs at glancing angles. The grounding 
operator $\mathcal{G}(\cdot)$ then snaps the bearing to a 
visibly safer sector and forces an unnecessary detour. 
Individually these detours are minor, but they prolong cue-free 
exposure and indirectly amplify failure mode (2). We observed 
this most often on the Scout Mini, whose more conservative 
$\tau_{\text{trav}}=0.7$ excludes terrain the platform could 
in fact traverse.

\textit{(4) Dynamic obstacles.}
Pedestrians, vehicles, or other agents entering the near field 
require sub-second reactions, but our VLM-based feasibility is 
refreshed only every 2--3 seconds. In real-world deployments, 
unexpected interference from pedestrians or other moving agents 
occasionally forced us to take over the robot for safety, 
ending the episode in failure.

\textbf{Future work.}
The failure modes above suggest three directions: tighter 
local planning for dynamic obstacle recovery (4), global priors 
such as GPS or OpenStreetMap to bound memory drift (2), and 
improved traversability estimation for ambiguous terrain (3).

\bibliographystyle{./bibliography/IEEEtran}
\bibliography{./bibliography/ref}
\clearpage

\end{document}